\pdfoutput=1
\documentclass[11pt]{article}

\usepackage{ACL2023}

\usepackage{times}
\usepackage{latexsym}
\usepackage{graphicx}
\usepackage{multirow}
\usepackage{multicol}
\usepackage{float}

\usepackage{caption}
\usepackage{subcaption}

\usepackage{colortbl}
\usepackage{xcolor}
\definecolor{blue}{HTML}{1A85FF}

\usepackage{booktabs}

\usepackage[T1]{fontenc}
\usepackage[utf8]{inputenc}

\usepackage{microtype}

\usepackage{inconsolata}

\newcommand{\datasetname}{\textsc{grotoap2-lds}}
\newcommand{\namelowdiversity}{\textsc{LimitedPublisher}}
\newcommand{\namemeddiversity}{\textsc{LimitedPublisher+}}
\newcommand{\namehighdiversity}{\textsc{LimitedPublisher++}}

\title{Are Layout-Infused Language Models Robust to Layout Distribution Shifts? A Case Study with Scientific Documents}

\author{Catherine Chen$^{\spadesuit\,\thanks{$^*$Work primarily done during internship at AI2.}}$ \quad Zejiang Shen$^{\circ}$ \quad Dan Klein$^\spadesuit$ \\  {\bf Gabriel Stanovsky}$^{\diamondsuit\heartsuit}$ \quad {\bf Doug Downey}$^{\diamondsuit\Box}$  \quad {\bf Kyle Lo}$^{\diamondsuit}$ \vspace{8pt}\\
$^\spadesuit$University of California Berkeley, $^\circ$Massachusetts Institute of Technology, \\$^\diamondsuit$Allen Institute for AI, $^\heartsuit$Hebrew University of Jerusalem, $^\Box$Northwestern University \\ \texttt{\small\{cathychen,klein\}@berkeley.edu} \quad \texttt{\small\{zjshen\}@mit.edu} \quad  \texttt{\small\{gabis,dougd,kylel\}@allenai.org} }

\begin{document}
\graphicspath{{./figures/}}
\maketitle
\begin{abstract}
Recent work has shown that infusing layout features into language models (LMs) improves processing of visually-rich documents such as scientific papers. Layout-infused LMs are often evaluated on documents with familiar layout features (e.g., papers from the same publisher), but in practice models encounter documents with unfamiliar distributions of layout features, such as new combinations of text sizes and styles, or new spatial configurations of textual elements. In this work, we test whether layout-infused LMs are robust to layout distribution shifts. As a case study, we use the task of scientific document structure recovery, segmenting a scientific paper into its structural categories (e.g., \textsc{title}, \textsc{caption}, \textsc{reference}). To emulate distribution shifts that occur in practice, we re-partition the \textsc{grotoap2} dataset. We find that under layout distribution shifts model performance degrades by up to 20 F1.  Simple training strategies, such as increasing training diversity, can reduce this degradation by over 35\% relative F1; however, models fail to reach in-distribution performance in any tested out-of-distribution conditions. This work highlights the need to consider layout distribution shifts during model evaluation, and presents a methodology for conducting such evaluations.\footnote{Our code and evaluation suite are available at \href{https://github.com/cchen23/layout_distribution_shift}{https://github.com/cchen23/layout\_distribution\_shift}.}
\end{abstract}

\section{Introduction}\label{sec:introduction}
Humans use layout to understand the organizational structure of visually-rich documents such as scientific papers, newspaper articles, and web pages. For instance, a reader might use fontsize and boldfacing to recognize a section title, while they might use spatial location to recognize a footnote. Based on the intuition that layout aids in document understanding, recent work introduced layout-infused language models (LMs). To improve document processing, these models incorporate layout features such as the styling, size, and spatial configuration of document text. However, these features often change between documents – are layout-infused LMs robust to shifts in layout distribution?

\begin{figure}[t!]
\centering
 	\includegraphics[width=\linewidth]{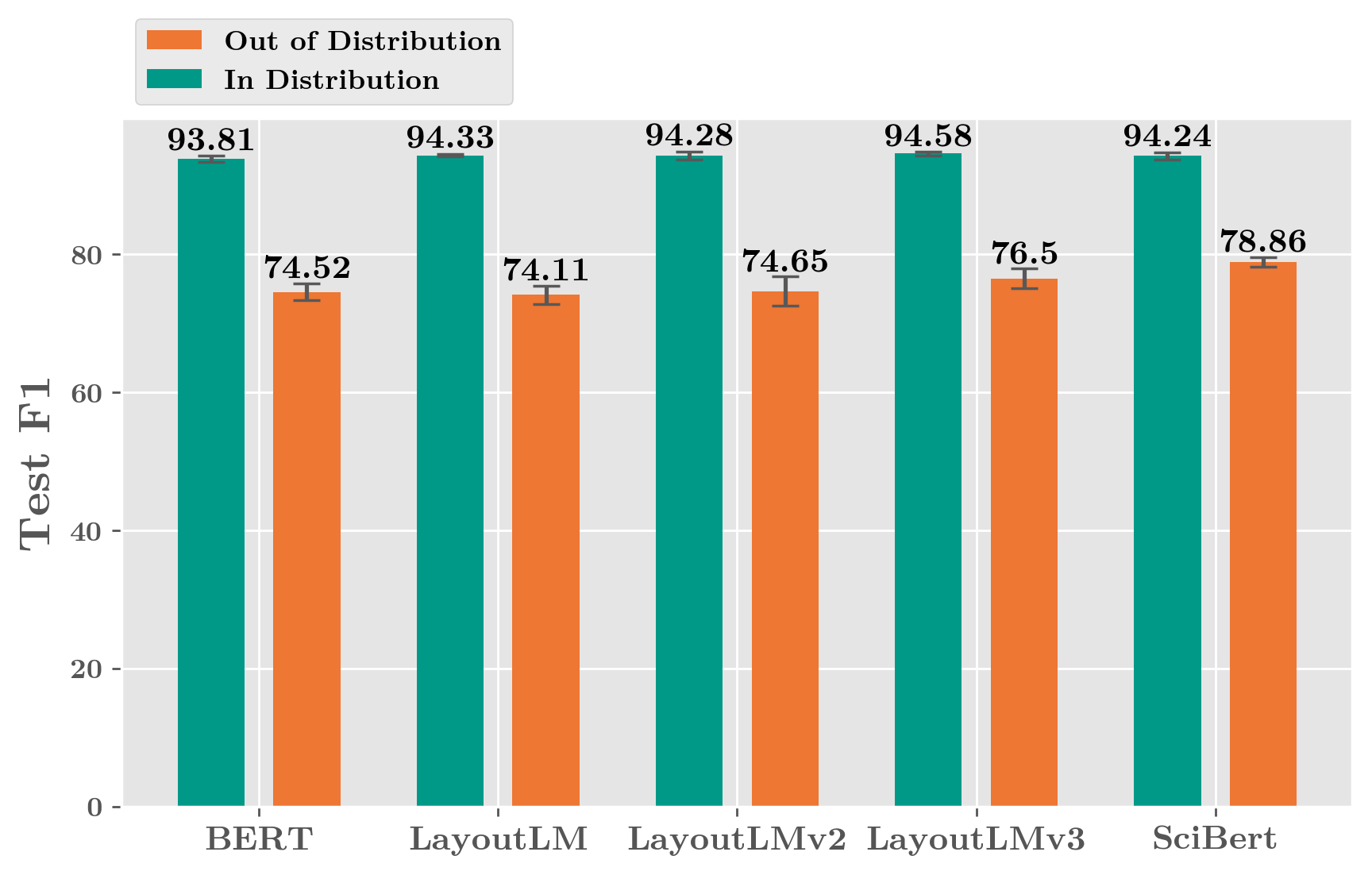}
\caption{Model performance on document structure recovery, comparing training and testing in-distribution vs out-of-distribution. Error bars indicate standard deviation across runs. Layout distribution shifts degrade model performance by up to 20 F1 (Section \ref{sec:methods_rq1}). Simple training strategies such as few-shot fine-tuning and increasing training diversity partially mitigate the drop shown here (Section \ref{sec:methods_rq2}, Section \ref{sec:methods_rq3}).}\label{fig:summary_figure}
\end{figure}

\begin{figure*}[t!]
\centering
 	\includegraphics[width=0.6\linewidth]{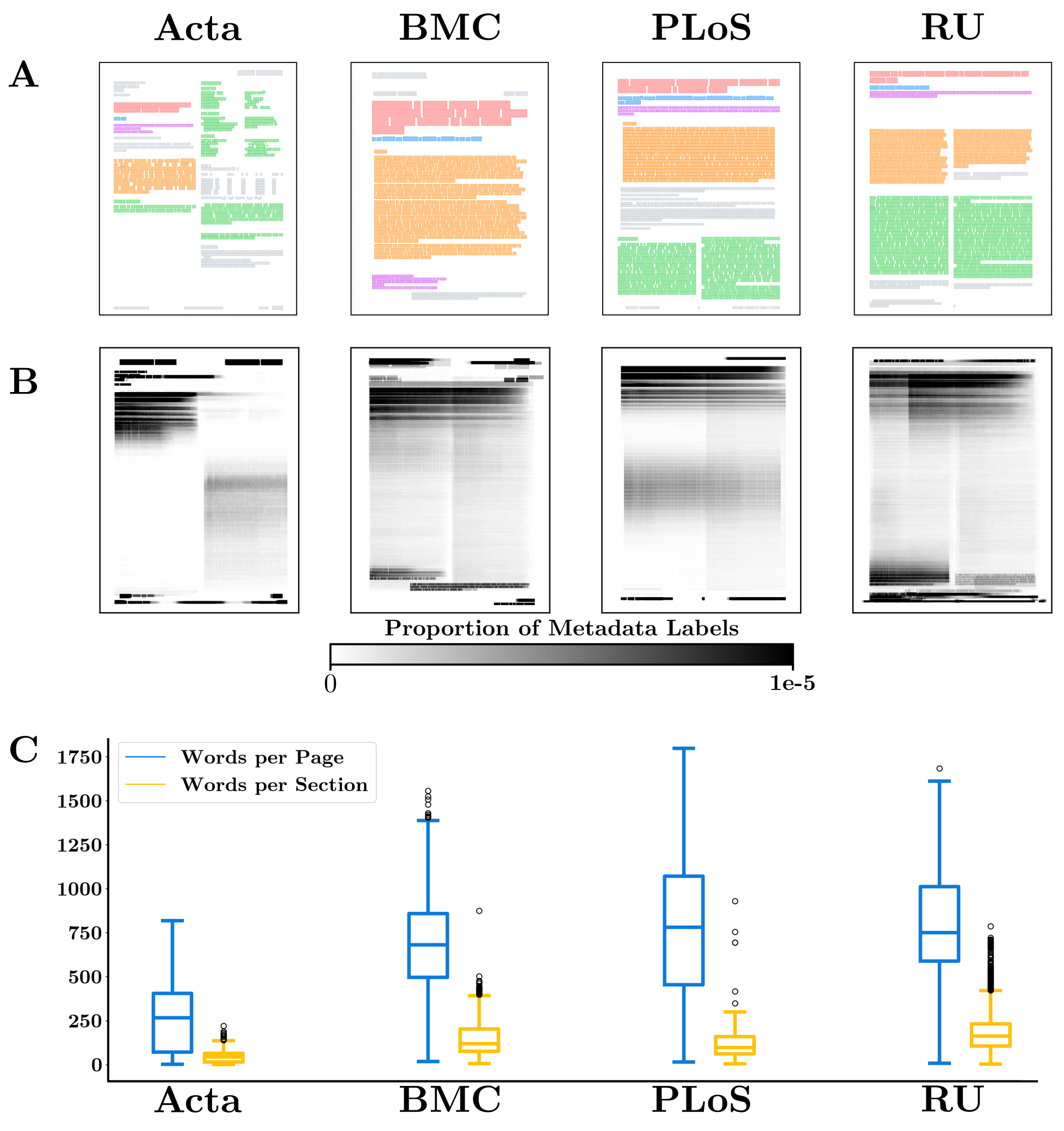}
\caption{Differences in layout between publishers (\textsc{Acta}, \textsc{BMC}, \textsc{PLoS}, and \textsc{RU}). \textbf{A}. One representative page from each publisher. Each word is colored according to structural category. Abstracts (orange) are structured in one (\textsc{Acta}, \textsc{BMC}, \textsc{PLoS}) or two (\textsc{RU}) columns, spanning half (\textsc{Acta}) or full (\textsc{BMC}, \textsc{PLoS}, \textsc{RU}) page widths. Author affiliations (pink) may appear at the top (\textsc{Acta}, \textsc{PLoS}, \textsc{RU}) or bottom (\textsc{BMC}) of the page. \textbf{B}. Spatial distribution of metadata categories (e.g., \textsc{author\_title}, \textsc{page\_num}). \textbf{C}. Boxplots showing number of words per page and section, for each publisher. }\label{fig:layout_variation_demo}
\end{figure*}

With rising interest in processing visually-rich documents, some language models have been augmented with components specifically designed to process layout features \cite[e.g.,][]{xu2020layoutlm}. Layout-infused models accurately process documents with similar layouts to those seen during training \citep{shen2022vila, huang2022layoutlmv3}, and can leverage visual information to better understand long-range dependencies \citep{Nguyen2023LoRaLayAM}. But in practice, models often encounter documents with different layouts – for instance, pages with a different number of columns, a different density of words on the page, and different locations of textual elements. In order to realistically evaluate model performance, we study model performance under layout distribution shifts. Although robustness to text-distribution shifts has been relatively well-studied, layout distribution shifts pose unique challenges and may require unique solutions; therefore, we focus specifically on robustness to layout distribution shifts.

Here we present a case study to evaluate model robustness to layout distribution shifts. Our case study focuses on the task of segmenting a scientific paper into its structural categories. For example, a scientific paper might be segmented into categories such as \textsc{title}, \textsc{authors}, \textsc{caption}, \textsc{paragraph}, and \textsc{reference}. We refer to this task as {\em document structure recovery}.\footnote{Other names for this task include layout analysis \citep{li2020docbank,zhong2019publaynet}, logical structure recovery \citep{huang-etal-2022-lightweight}, zone classification \citep{tkaczyk2014grotoap2}, and text classification \citep{shen2022vila}.} We chose to focus on this task because it serves as a testbed to determine whether layout-infusion remains beneficial under layout distribution shifts.  Document structure recovery requires grasping the organizational structure of a document, a key piece of information that layout conveys. Moreover, layout-infused models have been shown to reach state-of-the-art performance on this task, in settings where the layout distribution is the same between training and testing \citep{shen2022vila, huang2022layoutlmv3}. 

To test model robustness, evaluations must train and test on examples drawn from different layout distributions. However, existing datasets for document structure recovery use random train-test splits, and thus are ill-suited for evaluating model robustness. In this work we leverage publisher metadata to construct train-test splits that reflect layout distribution shifts. Publisher metadata is a proxy for layout distribution shifts because publication venue is a key driver of layout differences -- different publishers adhere to different style guides and templates (e.g., Figure \ref{fig:layout_variation_demo}). Moreover, layout differences across publishers reflect layout distribution shifts faced in practice as new publishers, templates, and style guides arise. We use publisher metadata to propose new train-test splits of an existing dataset (\textsc{grotoap2}, \citet{tkaczyk2014grotoap2}) for scientific document structure recovery. These splits are designed to reflect \textsc{\textbf{l}}ayout \textsc{\textbf{d}}istribution \textsc{\textbf{s}}hifts; hence we refer to the splits as \datasetname.

Using \datasetname, we evaluate a set of layout-infused LMs and find that model performance degrades by up to 20 F1 under layout distribution shifts (Figure \ref{fig:summary_figure}). We show that layout-infused models can quickly adapt to new distributions, and that increasing training diversity can improve model robustness. However, even with diverse training sets and fewshot fine-tuning, performance on out-of-distribution layouts remains more than 2 F1 below in-distribution performance. Although layout-infusion aids in processing documents with in-distribution layouts, layout-infused LMs may overfit on features seen during training. We release our code and evaluation suite to enable future evaluations and to facilitate expansions of our evaluation suite.

\section{Background and Related Work}
Recent work established that models are often sensitive to distribution shifts. Shifts in the distribution of text or image statistics have been shown to substantially degrade model performance \cite[e.g.,][]{geirhos2020shortcut,Bai2021PretrainOA,Ye2021CrossFitAF,Miller2020TheEO,koh2021wilds}, even in cases when human performance is robust to these distribution shifts \citep{Miller2020TheEO}. For example, question-answering models struggle to generalize from text in Wikipedia to text in newspaper articles \citep{Miller2020TheEO}, and image classification models struggle to generalize between images taken from different cameras \citep{koh2021wilds}.  We extend this line of research to study robustness to shifts in the distribution of layout features.

Document structure recovery provides an opportune setting for evaluating robustness to layout distribution shifts. Solving this task requires understanding how the text and visual layout of a page convey the organizational structure of the document, and layout-infused LMs have been shown to reach near-human performance on this task \citep{tkaczyk2014grotoap2,shen2022vila}. Prior work has shown that models transfer poorly across different document types (e.g., from scientific papers to financial documents) \citep{pfitzmann2022doclaynet}. Although different document types exhibit differences in layout, they also exhibit large differences in other features, such as the textual domain and the distribution of structural categories. It is therefore unclear whether poor transfer across document types is due to layout distribution shifts or other factors. In this work, we experiment using train-test splits exhibiting different layout distributions but with documents of the same type (i.e., scientific papers from biomedical journals).\footnote{Concurrent work \cite{Wang2022ABF} evaluates how well models extract information on unseen form types (e.g., training on ``Amendment" and ``Short Form" foreign agent registration forms from the US government, and testing on ``Dissemination Report" forms). In contrast, our experiments evaluate transfer on documents of the same type (scientific articles from biomedical journals), and experiment on a wider variety of layout transfer settings (e.g., varying the amount of  layout diversity seen during training) (Section \ref{sec:methods_rq3}).}

Existing evaluation datasets for document structure recovery include many document types, such as scientific papers \citep{tkaczyk2014grotoap2,li2020docbank,zhong2019publaynet}, forms \citep{jaume2019funsd}, receipts \citep{park2019cord}, and long-form business documents \citep{Gralinski2020KleisterAN,pfitzmann2022doclaynet}. We focus on scientific papers, where layout distribution shifts are prevalent (Figure \ref{fig:layout_variation_demo}). Although existing datasets for scientific document structure recovery contain documents with different layouts, existing train-test splits do not reflect layout distribution shifts.

\section{Evaluation Methodology}\label{sec:methods}
To evaluate model robustness, we propose a set of new train-test splits of \textsc{grotoap2}. These splits reflect layout distribution shifts that occur in practice, and we refer to this set of splits as \datasetname. In this section we formally define our task (Section \ref{sec:task_definition}), describe our procedure for partitioning data into splits that emulate layout distribution shifts (Section \ref{sec:dataset_construction_procedure}), and present a specific benchmark for evaluating robustness to layout distribution shifts (\datasetname, Section \ref{sec:dataset}).

\subsection{Task Definition}\label{sec:task_definition}
For each page, a model receives $N$ words $w_0,...,w_N$ in detected reading order. Layout-infused models receive additional page features, such as the x- and y- coordinates of the bounding box of each word or an image of the page. Given these inputs, the model must predict category labels $y_0,...,y_N$, one for each word, where $y_i$ is selected from a set of structural page categories (e.g., \textsc{title}, \textsc{caption}, \textsc{authors})

\subsection{Dataset Construction Procedure}\label{sec:dataset_construction_procedure}
We focus on layout distribution shifts within scientific papers, but our data partitioning procedure is agnostic to the particular document type. In the future, this procedure could be used to evaluate model robustness with other types of documents, such as receipts from different vendors or articles from different newspapers.
\paragraph{Document-Level Layout Assignments}
Previous evaluation setups assigned dataset splits at the page level, sometimes placing different pages of the same document in both the train and the test set~\citep{tkaczyk2014grotoap2,li2020docbank}. However, layout formatting decisions are often made at the document-level, and the layout of different pages in a multi-page document are often highly dependent on each other. We therefore consider layout in terms of whole documents, and assign dataset splits at the document level.
\paragraph{Provenance Metadata as a Proxy for Layout Distribution Shifts}
For scientific papers, different publishers format papers with different layouts \citep{tkaczyk2014grotoap2}, and layout differences across publishers reflect distribution shifts that may occur in practice. We therefore use publisher metadata to partition documents into different dataset splits.
Existing datasets often do not preserve provenance metadata, instead including only the content and task labels for each document \cite[e.g.,][]{tkaczyk2014grotoap2,li2020docbank,zhong2019publaynet}. Fortunately, scientific literature citation tools provide a way to recover publisher metadata for scientific papers. To link each paper to its associated publisher, we access the Semantic Scholar database to obtain the journal based on the title of each publication in \textsc{grotoap2}, and then map from each journal to the corresponding publisher.

\subsection{The \datasetname~Benchmark}\label{sec:dataset}
We use the procedure described in Section \ref{sec:dataset_construction_procedure} to construct \datasetname, a set of train-test splits that evaluate model performance under different training conditions.
\paragraph{Test splits}\label{sec:testsplits}
We construct four test sets, each of which contains papers from a held-out publisher (\textsc{Acta}, \textsc{BMC}, \textsc{PLoS}, \textsc{RU}). \textsc{grotoap2} contains a large number of papers from each of these publishers (at least 300 papers per publisher, or about 1 million words), ensuring enough data to compute reliable estimates of in-distribution and out-of-distribution performance.\footnote{Other publishers also met the minimum number of papers/words (e.g., Nucleic Acids Research). We focused on four publishers to keep the number of experiments tractable.}  Each of these four publishers contains papers from a qualitatively distinct layout distribution (Figure \ref{fig:layout_variation_demo}). The same four test sets are used to evaluate models under each train condition.

For each publisher, 20\% of papers were used as a held-out test set, and the remaining 80\% of papers were used in certain training conditions (e.g., to compute an estimate of in-distribution performance). The test sets contain an average of 75 papers ($\approx$500,000 words) each.

\datasetname~includes 12 label categories (\textsc{abstract}, \textsc{acknowledgements}, \textsc{affiliation}, \textsc{author\_title}, \textsc{bibliographic\_info}, \textsc{body}, \textsc{dates}, \textsc{figure}, \textsc{page\_num}, \textsc{references}, \textsc{table}, \textsc{unknown}). Train/test split sizes are included in Table~\ref{tab:grotoap2_lds_train_test_split_sizes}.

\begin{table}[!t]
\small
\centering
\begin{tabular}{lcc}
\specialrule{2pt}{1pt}{1pt}
Test publisher & Train papers & Test papers \\
\specialrule{1pt}{1pt}{1pt}
\textsc{Acta}\vspace{0.02em} & $2039$ & $44$\vspace{0.02em}\\
 \hline
\textsc{BMC}\vspace{0.02em} & $1886$ & $63$\vspace{0.02em}\\
 \hline
\textsc{RU}\vspace{0.02em} & $1893$ & $62$\vspace{0.02em}\\
 \hline
\textsc{PLoS}\vspace{0.02em} & $1349$ & $130$\vspace{0.02em}\\
\specialrule{2pt}{1pt}{1pt}
\end{tabular}
\caption{\datasetname~split sizes.}\label{tab:grotoap2_lds_train_test_split_sizes}
\end{table}
\paragraph{In-Distribution (ID) Training}
For each of the four held-out test publishers, we construct a training set with papers from the same publisher.  Papers from the same publisher exhibit different layouts, but layout differences between papers within the same publisher are small relative to differences between papers from different publishers. We therefore refer to settings in which models are trained and tested on papers from the same publisher as the “in-distribution” setting, and settings involving transfer across publishers as the “out-of-distribution” setting. Model performance in this setting is used to estimate the performance drop between in-distribution and out-of-distribution layouts.

\paragraph{Out-of-Distribution (OOD) Training}\label{sec:ood_datasets}
We construct training sets that evaluate model performance under layout distribution shift. The number of train papers is matched between training sets. Each training set contains roughly 2,000 papers ($\approx$10,000,000 words). We construct training sets reflecting different levels of layout diversity. Our default training approach (``\namelowdiversity”) is a leave-one-publisher-out setting in which each model is trained on three publishers and tested on the held-out fourth publisher. To evaluate the impact of training set diversity on robustness to layout distribution shifts, we construct datasets with 25 publishers (``\namemeddiversity”) or 125 publishers (``\namehighdiversity”).

To quantify the diversity of spatial configurations in each training set, we measure the breadth ($B$) of spatial locations covered by each structural category. To compute $B$, for each structural category we count the proportion of spatial x-y positions where that category occurs,\footnote{x-y positions are determined using pixel locations in images of each page.} and then compute the mean across categories.  The value of $B$ for each data split is included in Table \ref{tab:dataset_description}.
\begin{table}[!t]
\small
\centering
\begin{tabular}{lc}
\specialrule{2pt}{1pt}{1pt}
Data split & $B$ (\%) \\
\specialrule{1pt}{1pt}{1pt}
In-Distribution\vspace{0.02em} & $50.13$ \vspace{0.02em}\\
 \hline
\namelowdiversity\vspace{0.02em} & $64.54$\vspace{0.02em}\\
\hline
\namemeddiversity\vspace{0.02em} & $75.47$\vspace{0.02em}\\
\hline
\namehighdiversity\vspace{0.02em} & $78.46$\vspace{0.02em}\\
\specialrule{2pt}{1pt}{1pt}
\end{tabular}
\caption{Breadth of layouts in each training set. $B$ denotes the percentage of spatial page locations covered by each structural category, averaged over categories.}\label{tab:dataset_description}
\end{table}

\paragraph{Few-shot Adaptation} In practice, it may be possible to cheaply annotate a few papers from a new layout distribution (e.g., when a trained model is applied to papers from a new publisher.) To test how quickly models can adapt to a new layout distribution, we additionally evaluate models in settings in which models are first trained on an out-of-distribution training set, and are then fine-tuned on a small amount of in-distribution data. Specifically, before testing models on each of the test sets, we perform few-shot fine-tuning with a few annotated examples (10 papers, $\approx 50,000$ words) from the held-out test publisher.

\section{Experiment Details}
\subsection{Models}

We evaluate on BERT, LayoutLM, LayoutLMv2, LayoutLMv3, and SciBERT (we use the base uncased version of each model).\footnote{Because of computational constraints we evaluate on a subset of all existing layout-infused models. We release our code and train-test splits to aid evaluations of other models.} The three layout-infused models (LayoutLM, LayoutLMv2, LayoutLMv3) share the same model size and underlying architecture as BERT \cite{devlin2019bert}. The equivalence in model size facilitates direct comparisons between different methods of incorporating layout features. Each layout-infused model is adapted to use layout features such as text position or page image embeddings on top of the standard BERT architecture. These layout-infused models have previously been shown to achieve state-of-the-art performance for processing visually-rich text documents with in-distribution layouts \citep{xu2020layoutlm,xu-etal-2021-layoutlmv2,huang2022layoutlmv3,shen2022vila}. We briefly describe these layout-infused models, and defer to the original papers for more specific details about model architecture and training.
    
	LayoutLM \citep{xu2020layoutlm}: LayoutLM is initialized from BERT, and then adapted to incorporate information about spatial text position. Masked visual-language modeling and multi-label document classification are used to adapt the model to incorporate the layout-specific components.

	LayoutLMv2 \citep{xu-etal-2021-layoutlmv2}: LayoutLMv2 is initialized from BERT, and then adapted to incorporate spatial text position as well as image embeddings of page regions. Masked visual-language modeling and text-image alignment are used to adapt the model to incorporate the layout-specific components.

    LayoutLMv3 \citep{huang2022layoutlmv3}: LayoutLMv3 is initialized from RoBERTa, and then adapted to incorporate spatial text position as well as image embeddings of page patches. Masked language modeling, masked image modeling, and word-patch alignment are used to adapt the model to incorporate the layout-specific components.

We additionally evaluate on SciBERT \cite{Beltagy2019SciBERT}, which is pretrained with the same pretraining tasks as BERT, but instead with data from scientific texts. SciBERT allows us to compare the benefit of layout-infusion with the benefit of simply using a model pretrained on in-domain text.

I-VILA tokens, which provide a textual indication of visual group boundaries as part of model input, have been shown to improve performance on document structure recovery \citep{shen2022vila}. Our preliminary experiments showed that I-VILA tokens improve performance across all experimental  settings. Therefore for all reported experiments, we use block-level I-VILA tokens provided by \citet{shen2022vila}.

\subsection{Implementation Details}
We implemented experiments in PyTorch, using the transformers library to access pretrained models \cite{pytorch,huggingface}. The learning rate for each model was selected by training each model with a learning rate of 1e-04, 1e-05, and 1e-06, and selecting the learning rate with the best dev set performance. This learning rate sweep was done separately for the initial training phase and for few-shot fine-tuning (see Appendix for details). The initial training stage included a linear warmup schedule with 2000 steps. The adamW optimizer with $\beta_1=0.9, \beta_2=0.999$ was used during training. During each episode of few-shot fine-tuning, eight papers were used as the train set and two papers were used as the dev set. No warmup was used during few-shot fine-tuning. Batch size four was used throughout training. Models were trained for a maximum of 10 epochs during the initial training phase and a maximum of 250 epochs during few-shot fine-tuning. Dev set performance was used for early stopping. For each model, the initial training phase was performed over three random seeds. For each random seed, few-shot fine-tuning was performed over three different episodes. Tables in Section \ref{sec:results} show the mean and standard deviation over random seeds and episodes. Full experimental results are included in the Appendix.
\begin{table}[!b]
\small
\centering
\begin{tabular}{lccccc}
\specialrule{2pt}{1pt}{1pt}
Test split & Base model   & ID \\
\specialrule{1pt}{1pt}{1pt}
 \multirow{5}{*}{\textbf{\textsc{Split avg}}} & BERT     	& $93.81 $ \tiny $ \pm 0.42$      	\\
           	& LayoutLM 	& $94.33 $ \tiny $ \pm 0.18$      	\\
           	& LayoutLMv2   & $94.28 $ \tiny $ \pm 0.57$      	\\
           	& LayoutLMv3   & \cellcolor{blue!15} $\mathbf{94.58 }$ \tiny $  \pm 0.32$ \\
           	& SciBERT  	& $94.24 $ \tiny $ \pm 0.52$      	\\
\specialrule{1pt}{1pt}{1pt}
\multirow{5}{*}{ \textsc{Acta}} &  BERT &  $86.78 $ \tiny $ \pm 0.84$  \\
              	&  LayoutLM 	&  $86.71 $ \tiny $ \pm 0.23$            	\\
              	&  LayoutLMv2   &   $87.40$ \tiny $  \pm 0.69$ \\
              	&  LayoutLMv3   & $ \mathbf{88.67}$ \tiny $ \pm 0.34$        	\\
              	&  SciBERT  	&  $87.58 $ \tiny $ \pm 1.12$           	\\
\hline
\multirow{5}{*}{ \textsc{BMC}} &  BERT &  $95.82 $ \tiny $ \pm 0.41$   \\
              	&  LayoutLM 	&  $96.14 $ \tiny $ \pm 0.09$          	\\
              	&  LayoutLMv2   &  $95.77$ \tiny $  \pm 1.09$  \\
              	&  LayoutLMv3   &  $95.79 $ \tiny $ \pm0.63$                	\\
              	&  SciBERT  	& $\mathbf{96.23}$ \tiny $ \pm0.19$	\\
\hline
\multirow{5}{*}{ \textsc{RU}} &  BERT &  $95.38 $ \tiny $ \pm0.32$      	\\
              	&  LayoutLM 	&  $\mathbf{96.64 }$\tiny $ \pm0.20$        	\\
              	&  LayoutLMv2   &  $96.55$ \tiny $  \pm 0.17$   \\
              	&  LayoutLMv3   &  $96.36 $ \tiny $ \pm0.09$   \\
              	&  SciBERT  	&  $95.85 $ \tiny $ \pm0.37$          	\\
              	\hline
\multirow{5}{*}{ \textsc{PLoS}} &  BERT &  $97.24 $ \tiny $ \pm0.12$             	\\
              	&  LayoutLM 	&  $\mathbf{97.84} $ \tiny $ \pm0.19$    	\\
              	&  LayoutLMv2   &  $97.38 $ \tiny $ \pm 0.32$             	\\
              	&  LayoutLMv3   &  $97.48$ \tiny $  \pm 0.22$ \\
              	&  SciBERT  	&  $97.30 $ \tiny $ \pm0.42$     	\\
\specialrule{2pt}{1pt}{1pt}
\end{tabular}
\caption{\textbf{In-distribution performance.} Test macro-F1 on document structure recovery. Mean and standard deviation over trials is reported. The best performance is highlighted in blue. Layout-infused models achieve the highest in-distribution test performance.}\label{tab:publisher_transfer_in_domain}
\end{table}

\section{Results}\label{sec:results}

We use \datasetname~(Section \ref{sec:dataset}) to evaluate robustness to layout distribution shifts. For each experimental condition, models are evaluated on four test sets, each containing papers from a held-out layout distribution. Unless indicated otherwise, model performance is reported as the average across these four test sets.

\subsection{Layout-infused LMs Perform Best on In-Distribution Layouts}
In-distribution performance of each model is shown in Table \ref{tab:publisher_transfer_in_domain}. Consistent with prior work, we find that layout-infused LMs reach the highest performance for documents with in-distribution layouts.\footnote{Note that the inference-time costs of LayoutLMv2 and LayoutLMv3 are around $10\times$ more than other tested models.} In subsequent sections, we use $\Delta_{ID}$ to refer to the difference between model performance on this in-distribution training condition, and on out-of-distribution training conditions.

\subsection{Models Overfit to Layout Distributions Seen During Training}\label{sec:methods_rq1}

To evaluate model robustness to layout distribution shifts, we train models on papers from three publishers (\namelowdiversity), and then test on papers from a held-out test publisher. Model performance for each of the test sets is shown in Table \ref{tab:publisher_transfer_low_diversity_zeroshot}. Compared to in-distribution performance (Table \ref{tab:publisher_transfer_in_domain}), out-of-distribution performance drops between 15.38 and 20.22 F1 ($\Delta_{ID}$). Layout-infused models perform worse than SciBERT, a model not pretrained with layout-specific components. Although layout-infused models achieve the highest performance for in-distribution layouts, these models overfit to layout distributions seen during training. In settings in which models need to generalize to out-of-distribution layouts, models with in-distribution text pretraining (as with SciBERT) may be more effective.

\begin{table}[!t]
\small
\centering
\begin{tabular}{lccccc}
\specialrule{2pt}{1pt}{1pt}
Test split & Base model   & OOD          	& $\Delta_{ID}$ \\
\specialrule{1pt}{1pt}{1pt}
 \multirow{5}{*}{\textbf{\textsc{Split avg}}} & BERT     	& $74.52 $ \tiny $ \pm 1.22$      	& $-19.29 $      	\\
           	& LayoutLM 	& $74.11 $ \tiny $ \pm 1.30$       	& $-20.22 $ \tiny	\\
           	& LayoutLMv2   & $74.65 $ \tiny $ \pm 2.09$      	& $-19.63 $	\\
           	& LayoutLMv3   & $76.50 $ \tiny $ \pm 1.39$       	& $-18.07 $ \tiny  	\\
           	& SciBERT  	& \cellcolor{blue!15} $\mathbf{78.86 }$ \tiny $  \pm 0.74$ & \cellcolor{blue!15} $\mathbf{-15.38 }$ \tiny \\
\specialrule{1pt}{1pt}{1pt}
\multirow{5}{*}{ \textsc{Acta}} &  BERT &  $51.89 $ \tiny $ \pm 1.20$       	&  $-34.89$     	\\
              	&  LayoutLM 	&  $51.15 $ \tiny $ \pm 1.20$       	&  $-35.56$     	\\
              	&  LayoutLMv2   &  $55.83 $ \tiny $ \pm 0.98$      	&  $-31.57$     	\\
              	&  LayoutLMv3   &  $55.83 $ \tiny $ \pm 3.21$      	&  $-32.84$     	\\
              	&  SciBERT  	&   $\mathbf{60.66 }$ \tiny $  \pm 0.28$ &	$\mathbf{-26.92}$ \\
\hline
\multirow{5}{*}{ \textsc{BMC}} &  BERT    	&  $72.74 $ \tiny $ \pm 1.19$     	&  $-23.09$     	\\
              	&  LayoutLM 	&  $74.84 $ \tiny $ \pm 0.50$      	&  $-21.3$      	\\
              	&  LayoutLMv2   &  $73.26 $ \tiny $ \pm 1.86$     	&  $-22.51$     	\\
              	&  LayoutLMv3   &  $74.83 $ \tiny $ \pm 0.86$     	&  $-20.96$     	\\
              	&  SciBERT  	&	$\mathbf{78.10 }$ \tiny $  \pm 1.13$ &	$\mathbf{-18.13}$ \\
\hline
\multirow{5}{*}{ \textsc{RU}} &   BERT &  $83.68 $ \tiny $ \pm 2.02$      	&  $-11.7$      	\\
              	&  LayoutLM 	&  $82.91 $ \tiny $ \pm 1.52$      	&  $-13.73$     	\\
              	&  LayoutLMv2   &  $81.62 $ \tiny $ \pm 3.49$      	&  $-14.93$     	\\
              	&  LayoutLMv3   &  $84.13 $ \tiny $ \pm 1.24$      	&  $-12.23$     	\\
              	&  SciBERT  	&	$\mathbf{87.36 }$ \tiny $  \pm 0.79$ &	$\mathbf{-8.49}$  \\
\hline
\multirow{5}{*}{ \textsc{PLoS}} &  BERT &  $89.76 $ \tiny $ \pm 0.48$      	&  $-7.48$      	\\
              	&  LayoutLM 	&  $87.55 $ \tiny $ \pm 1.96$      	&  $-10.29$     	\\
              	&  LayoutLMv2   &  $87.87 $ \tiny $ \pm 2.03$      	&  $-9.51$      	\\
              	&  LayoutLMv3   &	$\mathbf{91.23 }$ \tiny $  \pm 0.23$ &	$\mathbf{-6.25}$  \\
              	&  SciBERT  	&  $89.32 $ \tiny $ \pm 0.76$      	&  $-7.98$      	\\
\specialrule{2pt}{1pt}{1pt}
\end{tabular}
\caption{\textbf{Out-of-distribution performance.} Test macro-F1 on document structure recovery. Layout distribution shift substantially degrades performance of all models, with SciBERT achieving the best out-of-distribution test performance.
For generalization to new layouts, in-domain text pretraining may be more effective than layout-infusion.
}\label{tab:publisher_transfer_low_diversity_zeroshot}
\end{table}

\begin{table}[!t]
\small
\centering
\begin{tabular}{lccccc}
\specialrule{2pt}{1pt}{1pt}
Test split & Base model   & OOD          	& $\Delta_{ID}$ \\
\specialrule{1pt}{1pt}{1pt}
 \multirow{5}{*}{\textbf{\textsc{Split avg}}} & BERT     	& $89.64 $ \tiny $ \pm 0.67$
      	& $-4.16 $    
  	\\
           	& LayoutLM 	& $90.14 $ \tiny $ \pm 0.69$
      	& $-4.19 $
  	\\
           	& LayoutLMv2   &  \cellcolor{blue!15}$\mathbf{90.95 }$ \tiny $  \pm 0.65$ &  \cellcolor{blue!15}$\mathbf{-3.32 }$      	\\
           	& LayoutLMv3   & $90.28 $ \tiny $ \pm 0.63$
      	& $-4.3 $
  	\\
           	& SciBERT  	& $90.50 $ \tiny $ \pm 0.58$
      	& $-3.73 $ \\
\specialrule{1pt}{1pt}{1pt}
\multirow{5}{*}{ \textsc{Acta}} &  BERT &  $79.03 $ \tiny $ \pm 0.93$      	&  $-7.75$	\\
              	&  LayoutLM 	&  $80.49 $ \tiny $ \pm 1.11$      	&  $-6.22$     	\\
              	&  LayoutLMv2   &   $\mathbf{81.45 }$ \tiny $  \pm 0.86$ &  $\mathbf{-5.95}$\\
              	&  LayoutLMv3   &  $79.89 $ \tiny $ \pm 0.85$      	&  $-8.78$   	\\
              	&  SciBERT  	&  $80.30 $ \tiny $ \pm 0.73$       	&  $-7.28$     	\\
\hline
\multirow{5}{*}{ \textsc{BMC}} &  BERT &  $92.10 $ \tiny $ \pm 1.02$      	&  $-3.73$     	\\
              	&  LayoutLM 	&  $93.32 $ \tiny $ \pm 0.73$     	&  $-2.82$       	\\
              	&  LayoutLMv2   &  $\mathbf{94.22 }$ \tiny $  \pm 0.70$ &  $\mathbf{-1.55}$ 	\\
              	&  LayoutLMv3   &  $93.19 $ \tiny $ \pm0.57$     	&  $-2.6$          	\\
              	&  SciBERT  	&  $93.74 $ \tiny $ \pm0.59$     	&  $-2.49$  	\\
\hline
\multirow{5}{*}{ \textsc{RU}} &  BERT &  $91.57 $ \tiny $ \pm0.19$      	&  $\mathbf{-3.81}$   	\\
              	&  LayoutLM 	&  $90.72 $ \tiny $ \pm0.41$      	&  $-5.92$  	\\
              	&  LayoutLMv2   &  $\mathbf{91.99 }$ \tiny $  \pm 0.47$ &  $-4.56$     	\\
              	&  LayoutLMv3   &  $91.64 $ \tiny $ \pm0.75$      	&  $-4.72$      	\\
              	&  SciBERT  	&  $91.79 $ \tiny $ \pm0.44$      	&  $-4.06$        	\\
              	\hline
\multirow{5}{*}{ \textsc{PLoS}} &  BERT &  $95.88 $ \tiny $ \pm0.54$      	&  $-1.36$        	\\
              	&  LayoutLM 	&  $96.04 $ \tiny $ \pm0.52$      	&  $-1.8$    	\\
              	&  LayoutLMv2   &  $96.15 $ \tiny $ \pm 0.58$      	&  $-1.23$           	\\
              	&  LayoutLMv3   &  $\mathbf{96.38 }$ \tiny $  \pm 0.36$ &  $\mathbf{-1.1}$    	\\
              	&  SciBERT  	&  $96.19 $ \tiny $ \pm0.56$      	&  $-1.11$       	\\
\specialrule{2pt}{1pt}{1pt}
\end{tabular}
\caption{\textbf{Out-of-distribution performance (test macro-F1), after few-shot adaptation.} Performance on OOD layouts falls below ID performance, but few-shot fine-tuning reduces the performance drop by up to 80\%. LayoutLMv2 achieves the best out-of-distribution test performance. Layout infusion may facilitate adaptation to new layout distributions.}\label{tab:publisher_transfer_low_diversity_fewshot}
\end{table}

\subsection{Models Can Quickly Adapt to Layout Distribution Shifts}\label{sec:methods_rq2}
In practice, it may sometimes be possible to cheaply annotate a few papers from a target distribution (e.g., when a system ingests papers from a new publisher). To test how well models can quickly adapt to a new layout distribution, we first train models on out-of-distribution layouts (\namelowdiversity). For each of the four test splits, we perform few-shot fine-tuning with ten papers from the held-out test publisher, and then evaluate on the test set for that publisher.

Table \ref{tab:publisher_transfer_low_diversity_fewshot} shows model performance in this setting. From the in-distribution to out-of-distribution settings, model performance drops between 3.3 and 4.3 F1 ($\Delta_{ID}$). Although model performance falls substantially below in-distribution performance, few-shot adaptation to the target distribution reduces the performance drop by over 80\% compared to settings in which models must directly generalize to the new distribution (Table \ref{tab:publisher_transfer_low_diversity_zeroshot}). After few-shot adaptation, LayoutLMv2 achieves the highest out-of-distribution test performance, suggesting that layout-infusion may help models adapt more quickly to new layout distributions.

\subsection{Increasing Layout Diversity Observed During Training Can Improve Robustness}\label{sec:methods_rq3}

 \begin{figure*}[!t]
	\centering
	\includegraphics[width=0.8\linewidth]{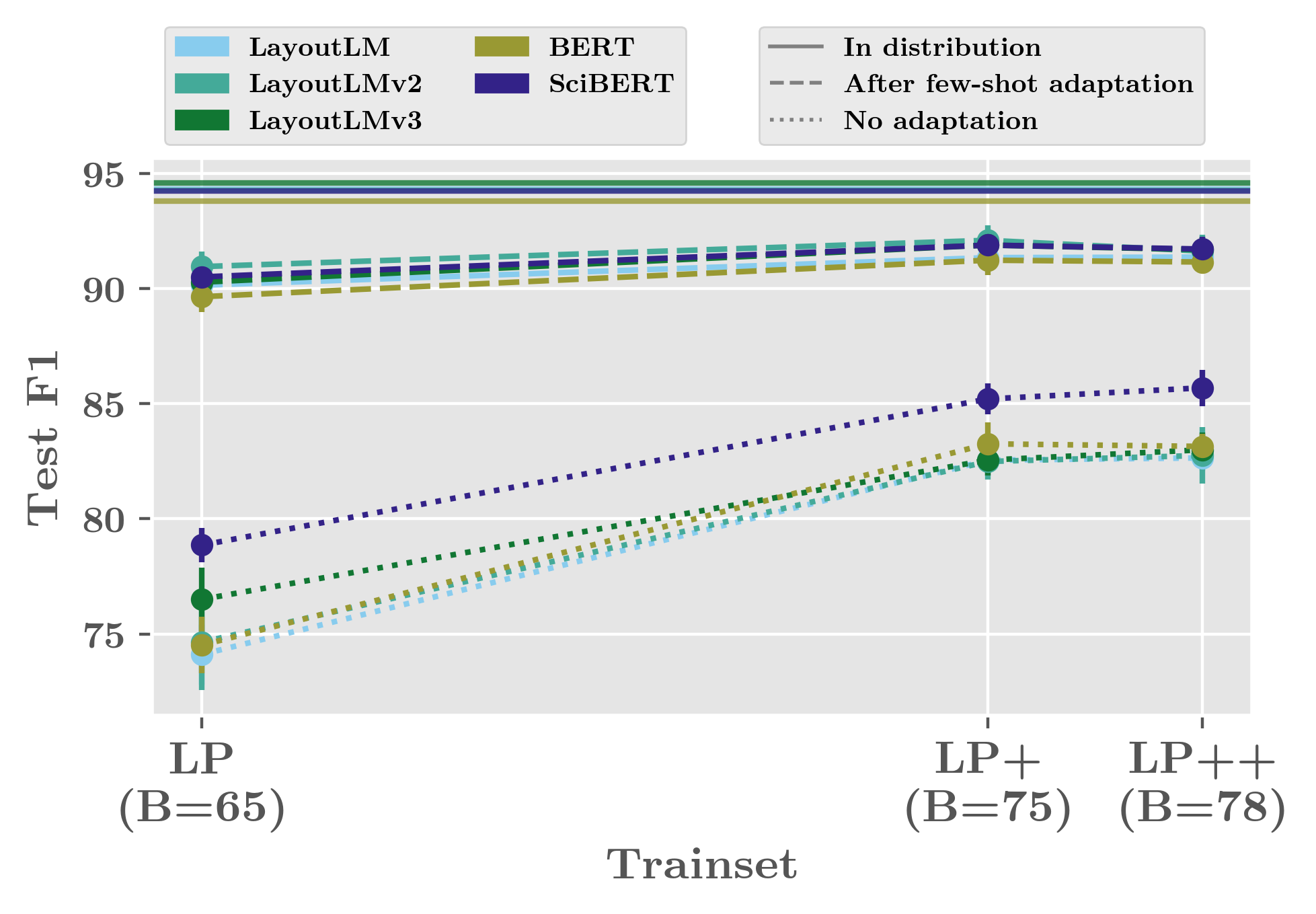}
	\caption{\textbf{Out-of-distribution performance vs training diversity.} Test macro-F1 on document structure recovery. LP=\textsc{\namelowdiversity}. Error bars reflect standard deviation over trials. (1) Increasing training diversity improves robustness to layout distribution shifts, but even the highest training diversity condition does not reach ID performance. (2) Increasing training diversity provides diminishing benefits. (3) Benefits of training diversity overlap with benefits from few-shot adaptation.}
\label{fig:layout_diversity_score_comparison}
\end{figure*}

To determine whether layout-diverse training can improve model robustness, we train models on papers from more publishers while holding the total number of papers constant (the \namemeddiversity~and \namehighdiversity~training sets described in Section \ref{sec:ood_datasets}). Model performance for each training diversity condition is shown in Figure \ref{fig:layout_diversity_score_comparison}. Performance is shown separately for settings in which models must generalize directly to papers from a different layout distribution (as in Section \ref{sec:methods_rq1}), and for settings in which models are fine-tuned on a few annotated examples from the target distribution (as in Section \ref{sec:methods_rq2}). 

When models must generalize directly to papers from a different layout distribution, a change from training on \namelowdiversity~to \namemeddiversity~increases test performance on out-of-distribution layouts by a mean of 9.91 F1 over models. A further increase in diversity from \namemeddiversity~to \namehighdiversity~increases performance by an additional 0.28 F1. In settings where models receive a few annotated examples to adapt to the target distribution (e.g., Section \ref{sec:methods_rq2}), training on \namemeddiversity~ rather than \namelowdiversity~yields a much smaller performance gain (1.53 F1). In few-shot adaptation settings, a further increase from training on \namemeddiversity~to \namehighdiversity~results in a -0.19 F1 drop in performance. 

These results suggest that increasing the diversity of layouts observed during training can improve model robustness, but that this strategy provides diminishing returns as training diversity continues to increase. Furthermore, the benefits of increasing training diversity may largely overlap with the benefits of few-shot adaptation to the target distribution. Even in the most favorable out-of-distribution setting, in which models are trained on the most beneficial training diversity condition and then fine-tuned on a few papers from the target layout distribution, model performance is at least 2 F1 below in-distribution performance.

\subsection{Error Analysis}\label{sec:error_analysis}
 \begin{figure}[!tb]
	\centering
	\includegraphics[width=\linewidth]{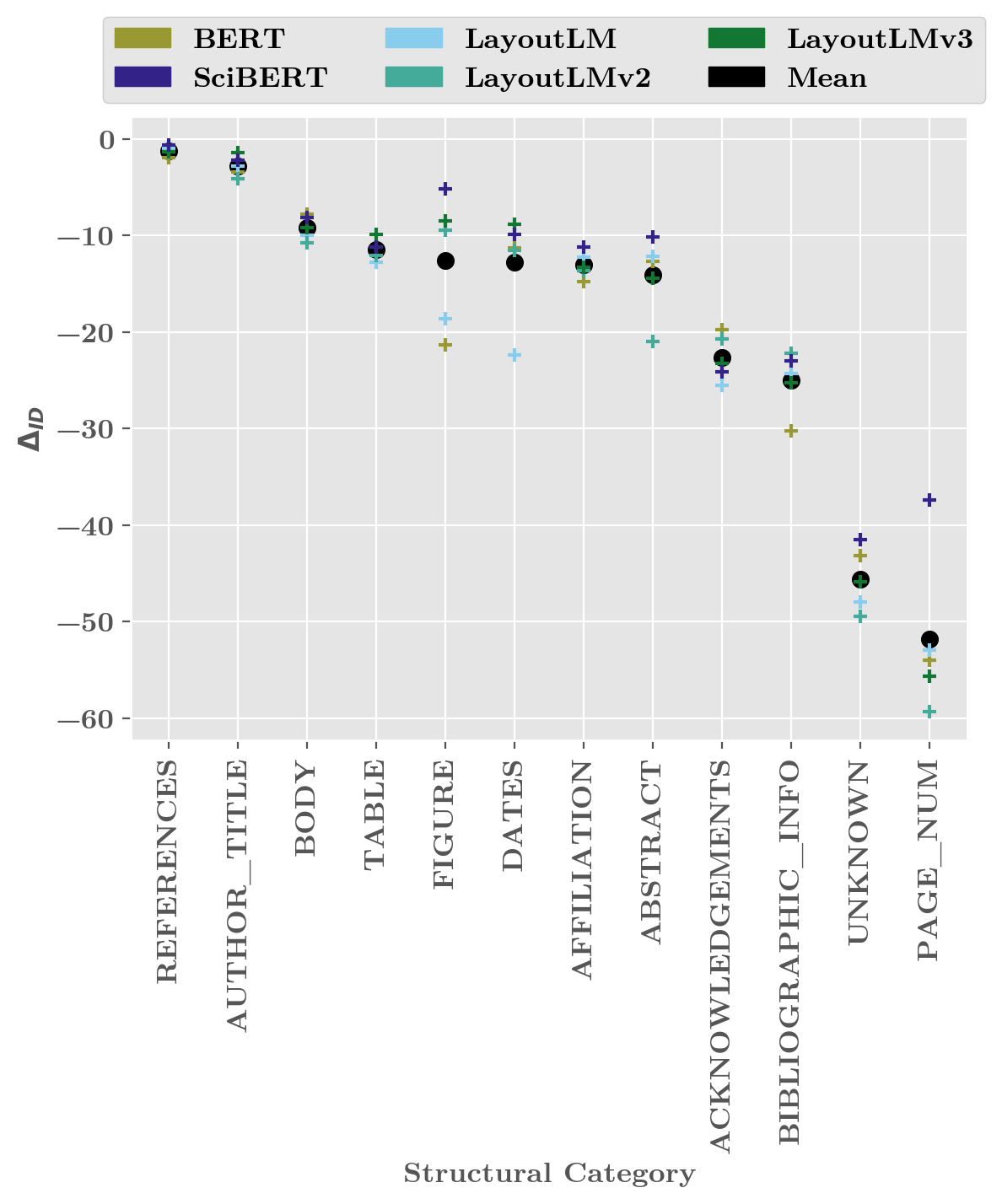}
	\caption{\textbf{Performance drop per category, low-diversity training.} Performance drop ($\Delta_{ID}$) between in-distribution and out-of-distribution layouts for each structural category. Largest performance drops occur for categories characterized by spatial page location (e.g., \textsc{page\_num}, \textsc{bibliographic\_info}). In contrast, much smaller performance drops occured in categories that contain the main textual content of the paper (e.g., \textsc{body}, \textsc{abstract})}
\label{fig:error_analysis_per_category}
\end{figure}

In practice, shifts in layout and text distribution are highly correlated. For instance, papers written for different scientific communities differ in both textual content and visual layout. To understand whether performance drops are driven by changes in layout, we analyzed model performance in the most difficult generalization setting (\namelowdiversity~with no few-shot adaptation). We examined whether generalization errors typically occurred for categories for which layout changes the most. Figure~\ref{fig:error_analysis_per_category} shows the performance drop between in-distribution and out-of-distribution settings for each structural category. Categories with the largest performance drops are those which are often characterized by spatial location, such as \textsc{page\_num} (-51.9 F1), \textsc{bibliographic\_info} (-25.0 F1), and \textsc{acknowledgements} (-22.7 F1). In contrast, much smaller performance drops occurred in categories containing the main textual content of the paper, such as \textsc{body} (-9.2 F1) and \textsc{abstract} (-14.1 F1).

\section{Conclusion}
This work studies whether layout-infused models are robust to layout distribution shift. We present a method for evaluating robustness to layout distribution shift, and construct \datasetname, a new set of splits for the \textsc{grotoap2} dataset that evaluate model robustness to layout distribution shifts. We use \datasetname~to evaluate a set of existing layout-infused models (LayoutLM, LayoutLMv2, and LayoutLMv3), and compare against two text-only models (BERT, SciBERT).

Layout-infused models perform most accurately on documents with familiar layouts (Table~\ref{tab:publisher_transfer_in_domain}), but in settings where models must generalize to documents with unfamiliar layouts, layout-infused models underperform text-only models such as SciBERT (Table~\ref{tab:publisher_transfer_low_diversity_zeroshot}). In such settings, models with in-domain text pretraining both provide more accurate results, and obviate the inference time cost of processing visual layout features (e.g., image embeddings LayoutLMv3 increase inference time by $\approx 10\times$).

We hypothesize that layout-specific components overfit more because they receive less pretraining data compared to text-only components, or because they increase total model parameter count (e.g., LayoutLM, LayoutLMv2, and LayoutLMv3 contain 20-45\% more parameters than BERT and SciBERT).\footnote{We note that the discrepancy in generalization performance is not driven by the proportion of UNK tokens. All tokenizers produced fewer than 0.3\% UNK tokens, and the tokenizer for SciBERT in fact had more UNK tokens than the tokenizer for BERT and the LayoutLM models.} Future work could test whether larger-scale pretraining improves robustness of layout-specific components.

We show that training strategies such as increasing training diversity or few-shot adaptation to the target layout distribution can mitigate the performance drop across layout distribution shifts. These results provide guidance for curating training data and highlight the importance during data collection of curating examples that reflect variation in document provenance. In situations with a known change in layout distribution (e.g., if a system trained on papers from one publisher is re-used to process papers from a new publisher), the cost of annotating a few examples from the target distribution may be highly effective, resulting in a large improvement in out-of-distribution model performance.

This work highlights the importance of considering layout distribution shifts when evaluating models on tasks involving visually-rich documents such as scientific papers. We hope that our study and evaluation methodology facilitate the development of layout-infused models that can generalize across layout distribution shifts.

\section{Limitations}
We use scientific papers as a first testbed for evaluating model robustness to layout distribution shifts. Many different layouts exist among scientific papers, and the existence of metadata databases facilitated the construction of train-test splits with layout distribution shifts. However, scientific papers are only one domain in which layout distribution shifts occur. Layouts also vary for many other visually-rich documents, such as business forms, receipts, webpages, and newspapers. We hope our evaluation methodology engenders evaluations on a wider range of document types.

Our experiments involve a subset of the many layout-infused models proposed in recent work \cite[e.g.,][]{Peng2022ERNIELayoutLK,Kim2021OCRFreeDU,Li2021SelfDocSD}. The models in our experiments were chosen because they share a similar model size and underlying architecture, facilitating comparisons between different methods of layout-infusion. We release our evaluation suite to enable more comprehensive evaluations in the future.

Performance drops occur both for layout-infused and, to a lesser extent, text-only models. The performance drops from text-only models may be due to layout information conveyed via word order and visual section boundary markers, but may also reflect shifts in text distribution. Our error analyses suggest that generalization errors are driven by shifts in layout rather than content (Section~\ref{sec:error_analysis}). In the future, synthetic experiments (e.g., with LaTeX-based manipulations) would help to fully disentangle the effects of layout and content distribution shifts, provided that large-scale synthetic manipulations can be constrained to produce realistic layouts.
\section{Potential Risks}
Although we do not foresee direct harms from this work, our work is related to automated processing of scientific documents. This line of study carries the risk of inaccurately processing documents and propagating false information about scientific findings.

\section{Acknowledgements}
This work was supported in part by NSF Grant 2033558. CC was supported in part by an IBM PhD Fellowship.

\bibliographystyle{acl_natbib}
\bibliography{references_fixed}

\newpage
\appendix
\section*{Appendix}
\subsection*{Experiment Compute Details}
All experiments were run on NVIDIA RTX A6000 GPUs. For each training condition, model training took around one day for LayoutLMv2 and LayoutLMv3, and took a couple hours for the other three models (LayoutLM, BERT, and SciBERT).

\subsection*{Learning Rate Selection}\label{sec:lr_selection}
To select the learning rate for each model, models were trained on each of three learning rates (1e-04, 1e-05, 1e-06), and the learning rate that produced the best performance on the dev set was selected. Learning rate selection was performed separately for each of the training stages: the initial stage of training on the larger set of out-of-distribution layouts, and few-shot fine-tuning on examples from the target distribution. Dev performance for each model is shown for the initial training stage (Table \ref{tab:lr_meta_train}) and few-shot fine-tuning (Table \ref{tab:lr_fewshot_finetune}).

\begin{table}[!h]
\centering
\begin{tabular}{lrrr}
\specialrule{2pt}{1pt}{1pt}
Base Model           	&   $1e-04$ &   $1e-05$ &   $1e-06$ \\
\specialrule{1pt}{1pt}{1pt}
 BERT    	&  $97.067$ &  $97.817$ &  $96.942$ \\
 LayoutLM	&  $93.964$ &  $98.066$ &  $97.308$ \\
 LayoutLMv2 &   $6.326$ &  $98.006$ &  $97.171$ \\
 LayoutLMv3   	&   $6.326$ &  $98.314$ &  $97.794$ \\
  SciBERT &  $96.644$ &  $98.184$ &  $97.902$ \\
\specialrule{2pt}{1pt}{1pt}
\end{tabular}
\caption{Learning rate selection for the initial training stage. Dev performance for each learning rate}\label{tab:lr_meta_train}
\end{table}

\begin{table}[!h]
\centering
\begin{tabular}{lrrr}
\specialrule{2pt}{1pt}{1pt}
  Base Model           	&   $1e-04$ &   $1e-05$ &   $1e-06$ \\
\specialrule{1pt}{1pt}{1pt}
 BERT    	&  $78.408$ &  $97.301$ &  $97.395$ \\
 LayoutLM	&  $79.399$ &  $97.685$ &  $97.839$ \\
 LayoutLMv2  &  $79.358$ &  $97.869$ &  $97.772$ \\
 LayoutLMv3      	&  $80.731$ &  $97.5$   &  $97.813$ \\
  SciBERT &  $77.417$ &  $97.733$ &  $98.01$  \\
\specialrule{2pt}{1pt}{1pt}
\end{tabular}
\caption{Learning rate selection for few-shot fine-tuning. Dev performance for each learning rate}\label{tab:lr_fewshot_finetune}
\end{table}

\subsection*{Dataset Details}
Our set of new train-test splits, \datasetname, is an adaptation of data released in the \textsc{grotoap2} dataset. \textsc{grotoap2} is distributed under the CC-BY license. Further use of \datasetname should attribute original dataset collection to \cite{tkaczyk2014grotoap2}. We refer the reader to \cite{tkaczyk2014grotoap2} for details of the original data collection procedure.

\subsubsection*{Label remapping}
Although a shared annotation procedure was used to label all papers in the \textsc{grotoap2} dataset \citep{tkaczyk2014grotoap2}, some differences XML formatting for different publishers resulted in discrepancies between structural category labels used for different publishers. For instance, in the original \textsc{grotoap2} dataset \textsc{title\_author} labels are used for some publishers, whereas separate \textsc{title} and \textsc{author} labels are used for other publishers). To account for minor annotation discrepancies between publishers as well as insufficient support for certain category labels in our dataset splits, we re-map the structural category tagset used in the original \textsc{grotoap2} dataset. Our label re-maping is shown in Table \ref{tab:remapping}.

\begin{table}[h]
\centering
\small
\begin{tabular}{p{2cm}p{4cm}}
\specialrule{2pt}{1pt}{1pt}
\textbf{New label}           	& \textbf{Original labels}\\
\specialrule{1pt}{1pt}{1pt}
\textsc{bib\_info} & \textsc{bib\_info}, \textsc{copyright} \\\hline
\textsc{title\_author} & \textsc{author}, \textsc{title}, \textsc{title\_author} \\\hline
\textsc{affiliation} & \textsc{affiliation}, \textsc{correspondence} \\\hline
\textsc{unknown} & \textsc{keywords}, \textsc{glossary}, \textsc{equation}, \textsc{type}, \textsc{editor}, \textsc{conflict\_statement}, \textsc{unknown} \\
\specialrule{2pt}{1pt}{1pt}
\end{tabular}
\caption{Label remapping for the \textsc{grotoap2} tagset.}\label{tab:remapping}
\end{table}

\subsection*{Full results}\label{sec:full_results}
We provide the test performance for each trial and episode in Tables \ref{tab:indistribution_zeroshot}, \ref{tab:lowdiversity_zeroshot}, \ref{tab:lowdiversity_fewshot}, \ref{tab:meddiversity_zeroshot}, \ref{tab:meddiversity_fewshot}, \ref{tab:highdiversity_zeroshot}, and \ref{tab:highdiversity_fewshot}.
\begin{table}
\tiny
\begin{tabular}{llrrr}
\specialrule{2pt}{1pt}{1pt}
Test Publisher   & Base Model   &   Seed 0 &   Seed 1 &   Seed 2 \\
\specialrule{1pt}{1pt}{1pt}
\multirow{5}{*}{\textsc{Acta}} & BERT         & $87.96$  & $86.05$  & $86.34$  \\
                  & LayoutLM     & $86.38$  & $86.82$  & $86.92$  \\
                  & LayoutLMv2   & $86.44$  & $88.02$  & $87.75$  \\
                  & LayoutLMv3   & $88.26$  & $88.63$  & $89.1$   \\
                  & SciBERT      & $86.02$  & $88.11$  & $88.59$  \\
\hline
\multirow{5}{*}{\textsc{BMC}} & BERT        & $95.66$  & $96.4$   & $95.44$  \\
                  & LayoutLM     & $96.13$  & $96.25$  & $96.03$  \\
                  & LayoutLMv2   & $96.5$   & $96.58$  & $94.24$  \\
                  & LayoutLMv3   & $94.94$  & $95.94$  & $96.48$  \\
                  & SciBERT      & $96.39$  & $96.34$  & $95.97$  \\
\hline
\multirow{5}{*}{\textsc{RU}} & BERT        & $95.49$  & $95.71$  & $94.94$  \\
                  & LayoutLM     & $96.66$  & $96.87$  & $96.39$  \\
                  & LayoutLMv2   & $96.68$  & $96.66$  & $96.3$   \\
                  & LayoutLMv3   & $96.29$  & $96.48$  & $96.3$   \\
                  & SciBERT      & $95.45$  & $95.75$  & $96.34$  \\
\hline
\multirow{5}{*}{\textsc{PLoS}} & BERT & $97.34$  & $97.07$  & $97.31$  \\
                  & LayoutLM     & $98.01$  & $97.92$  & $97.58$  \\
                  & LayoutLMv2   & $97.44$  & $96.97$  & $97.74$  \\
                  & LayoutLMv3   & $97.77$  & $97.42$  & $97.24$  \\
                  & SciBERT      & $97.38$  & $97.77$  & $96.76$  \\
\specialrule{2pt}{1pt}{1pt}
\end{tabular}
\caption{Model performance for each random seed, in-distribution training}\label{tab:indistribution_zeroshot}
\end{table}

\begin{table}
\tiny
\begin{tabular}{llrrr}
\specialrule{2pt}{1pt}{1pt}
Test Publisher   & Base Model   &   Seed 0 &   Seed 1 &   Seed 2 \\
\specialrule{1pt}{1pt}{1pt}
\multirow{5}{*}{\textsc{Acta}} & BERT        & $51.46$  & $53.52$  & $50.69$  \\
                  & LayoutLM     & $52.85$  & $50.25$  & $50.36$  \\
                  & LayoutLMv2   & $57.09$  & $54.69$  & $55.7$   \\
                  & LayoutLMv3   & $59.87$  & $55.58$  & $52.02$  \\
                  & SciBERT      & $60.37$  & $60.59$  & $61.03$  \\
\hline
\multirow{5}{*}{\textsc{BMC}} & BERT         & $71.55$  & $74.37$  & $72.31$  \\
                  & LayoutLM     & $75.54$  & $74.42$  & $74.55$  \\
                  & LayoutLMv2   & $70.65$  & $74.28$  & $74.85$  \\
                  & LayoutLMv3   & $73.63$  & $75.23$  & $75.62$  \\
                  & SciBERT      & $78.47$  & $76.58$  & $79.27$  \\
\hline
\multirow{5}{*}{\textsc{RU}} & BERT      & $81.03$  & $85.93$  & $84.07$  \\
                  & LayoutLM     & $84.97$  & $81.34$  & $82.42$  \\
                  & LayoutLMv2   & $82.49$  & $85.4$   & $76.98$  \\
                  & LayoutLMv3   & $83.04$  & $85.87$  & $83.49$  \\
                  & SciBERT      & $88.36$  & $87.3$   & $86.44$  \\
\hline
\multirow{5}{*}{\textsc{PLoS}} & BERT  & $89.09$  & $90.2$   & $90.0$   \\
                  & LayoutLM     & $84.9$   & $88.17$  & $89.58$  \\
                  & LayoutLMv2   & $90.4$   & $87.76$  & $85.43$  \\
                  & LayoutLMv3   & $91.56$  & $91.09$  & $91.04$  \\
                  & SciBERT      & $88.25$  & $89.9$   & $89.8$   \\
\specialrule{2pt}{1pt}{1pt}
\end{tabular}
\caption{Model performance for each random seed, \namelowdiversity~training}\label{tab:lowdiversity_zeroshot}
\end{table}

\begin{table}
\tiny
\begin{tabular}{llrrr}
\specialrule{2pt}{1pt}{1pt}
Test Publisher   & Base Model   &   Seed 0 &   Seed 1 &   Seed 2 \\
\specialrule{1pt}{1pt}{1pt}
\multirow{5}{*}{\textsc{Acta}} & BERT     & $61.6$   & $57.36$  & $59.53$  \\
                  & LayoutLM     & $58.8$   & $62.31$  & $59.99$  \\
                  & LayoutLMv2   & $60.4$   & $59.15$  & $58.22$  \\
                  & LayoutLMv3   & $60.01$  & $55.43$  & $57.11$  \\
                  & SciBERT      & $62.62$  & $62.94$  & $64.25$  \\     
\hline
\multirow{5}{*}{\textsc{BMC}} & BERT    & $94.1$   & $94.14$  & $95.18$  \\
                  & LayoutLM     & $93.53$  & $93.71$  & $94.42$  \\
                  & LayoutLMv2   & $94.01$  & $93.26$  & $93.9$   \\
                  & LayoutLMv3   & $95.01$  & $95.32$  & $94.86$  \\
                  & SciBERT      & $94.67$  & $94.56$  & $93.81$  \\
\hline
\multirow{5}{*}{\textsc{RU}} & BERT    & $92.15$  & $90.85$  & $92.06$  \\
                  & LayoutLM     & $91.43$  & $91.32$  & $91.28$  \\
                  & LayoutLMv2   & $92.65$  & $90.96$  & $90.29$  \\
                  & LayoutLMv3   & $90.87$  & $91.08$  & $90.49$  \\
                  & SciBERT      & $92.52$  & $92.68$  & $92.26$  \\
\hline
\multirow{5}{*}{\textsc{PLoS}} & BERT    & $88.0$   & $87.94$  & $86.17$  \\
                  & LayoutLM     & $84.52$  & $83.59$  & $85.66$  \\
                  & LayoutLMv2   & $86.01$  & $84.43$  & $86.59$  \\
                  & LayoutLMv3   & $86.61$  & $86.52$  & $87.33$  \\
                  & SciBERT      & $89.09$  & $92.61$  & $90.44$  \\
\specialrule{2pt}{1pt}{1pt}
\end{tabular}
\caption{Model performance for each random seed, \namemeddiversity~training}\label{tab:meddiversity_zeroshot}
\end{table}

\begin{table}[!t]
\tiny
\begin{tabular}{llrrr}
\specialrule{2pt}{1pt}{1pt}
Test Publisher   & Base Model   &   Seed 0 &   Seed 1 &   Seed 2 \\
\specialrule{1pt}{1pt}{1pt}
\multirow{5}{*}{\textsc{Acta}} & BERT  & $59.59$  & $57.42$  & $59.48$  \\
                  & LayoutLM     & $60.48$  & $58.04$  & $57.37$  \\
                  & LayoutLMv2   & $61.31$  & $55.37$  & $56.47$  \\
                  & LayoutLMv3   & $54.78$  & $59.63$  & $56.45$  \\
                  & SciBERT      & $61.26$  & $64.0$   & $65.55$  \\       
\hline
\multirow{5}{*}{\textsc{BMC}} & BERT            & $94.82$  & $94.64$  & $95.16$  \\
                  & LayoutLM     & $94.12$  & $94.51$  & $93.7$   \\
                  & LayoutLMv2   & $93.28$  & $94.4$   & $93.65$  \\
                  & LayoutLMv3   & $94.44$  & $94.4$   & $94.03$  \\
                  & SciBERT      & $93.54$  & $94.75$  & $93.68$  \\
\hline
\multirow{5}{*}{\textsc{RU}} & BERT   & $92.44$  & $92.37$  & $92.57$  \\
                  & LayoutLM     & $93.09$  & $92.29$  & $92.67$  \\
                  & LayoutLMv2   & $92.87$  & $92.15$  & $91.68$  \\
                  & LayoutLMv3   & $93.31$  & $93.49$  & $93.06$  \\
                  & SciBERT      & $91.68$  & $91.91$  & $92.57$  \\
\hline
\multirow{5}{*}{\textsc{PLoS}} & BERT    & $86.62$  & $87.41$  & $85.24$  \\
                  & LayoutLM     & $85.13$  & $85.54$  & $84.8$   \\
                  & LayoutLMv2   & $88.57$  & $87.84$  & $85.45$  \\
                  & LayoutLMv3   & $87.95$  & $87.84$  & $86.62$  \\
                  & SciBERT      & $92.54$  & $93.69$  & $93.1$   \\
\specialrule{2pt}{1pt}{1pt}
\end{tabular}
\caption{Model performance for each random seed, \namehighdiversity~training}\label{tab:highdiversity_zeroshot}
\end{table}

\begin{table*}
\tiny
\begin{tabular}{lllllllllll}
\specialrule{2pt}{1pt}{1pt}
Test Publisher   & Base Model   &   Seed 0  Ep 0 & Seed 0  Ep 1 & Seed 0  Ep 2 &   Seed 1  Ep 0 & Seed 1  Ep 1 & Seed 1  Ep 2 &   Seed 2  Ep 0 & Seed 2  Ep 1 & Seed 2  Ep 2 \\
\specialrule{1pt}{1pt}{1pt}
\multirow{5}{*}{\textsc{Acta}} & BERT        & $79.22$ & $79.93$ & $80.12$ & $79.48$ & $77.56$ & $80.07$ & $78.95$ & $78.0$  & $77.9$  \\
                  & LayoutLM     & $82.48$ & $81.22$ & $80.35$ & $80.44$ & $79.23$ & $78.49$ & $81.48$ & $80.36$ & $80.35$ \\
                  & LayoutLMv2   & $81.46$ & $82.2$  & $80.92$ & $81.55$ & $80.87$ & $79.63$ & $82.11$ & $82.75$ & $81.52$ \\
                  & LayoutLMv3   & $81.49$ & $79.56$ & $80.22$ & $80.23$ & $79.77$ & $79.98$ & $79.96$ & $77.99$ & $79.76$ \\
                  & SciBERT      & $80.78$ & $81.44$ & $79.62$ & $79.6$  & $80.5$  & $79.88$ & $81.05$ & $79.14$ & $80.73$ \\
\hline
\multirow{5}{*}{\textsc{BMC}} & BERT         & $90.14$ & $90.78$ & $92.54$ & $92.84$ & $91.44$ & $93.11$ & $92.89$ & $92.06$ & $93.09$ \\
                  & LayoutLM     & $92.46$ & $93.07$ & $93.6$  & $94.51$ & $93.76$ & $94.15$ & $92.09$ & $92.91$ & $93.35$ \\
                  & LayoutLMv2   & $93.35$ & $93.21$ & $93.81$ & $95.58$ & $93.97$ & $94.71$ & $94.77$ & $94.2$  & $94.41$ \\
                  & LayoutLMv3   & $93.15$ & $92.53$ & $93.73$ & $93.48$ & $93.48$ & $94.13$ & $92.37$ & $92.49$ & $93.38$ \\
                  & SciBERT      & $94.36$ & $93.38$ & $94.44$ & $93.76$ & $93.06$ & $94.28$ & $93.44$ & $92.73$ & $94.23$ \\
\hline
\multirow{5}{*}{\textsc{RU}} & BERT     & $91.74$ & $91.77$ & $91.5$  & $91.29$ & $91.66$ & $91.62$ & $91.66$ & $91.69$ & $91.22$ \\
                  & LayoutLM     & $90.86$ & $90.0$  & $90.56$ & $91.28$ & $91.12$ & $91.01$ & $90.93$ & $90.18$ & $90.49$ \\
                  & LayoutLMv2   & $92.08$ & $92.62$ & $91.5$  & $91.96$ & $91.96$ & $92.44$ & $91.78$ & $92.48$ & $91.06$ \\
                  & LayoutLMv3   & $91.98$ & $92.01$ & $91.88$ & $90.96$ & $90.79$ & $90.28$ & $92.85$ & $92.02$ & $91.99$ \\
                  & SciBERT      & $92.02$ & $90.99$ & $91.18$ & $92.28$ & $91.89$ & $91.63$ & $92.44$ & $91.74$ & $91.92$ \\
\hline
\multirow{5}{*}{\textsc{PLoS}} & BERT      & $95.94$ & $95.42$ & $96.69$ & $96.15$ & $95.41$ & $96.52$ & $96.07$ & $94.86$ & $95.88$ \\
                  & LayoutLM     & $96.91$ & $96.18$ & $95.92$ & $96.08$ & $95.43$ & $96.05$ & $96.81$ & $95.56$ & $95.38$ \\
                  & LayoutLMv2   & $97.12$ & $96.07$ & $95.93$ & $96.63$ & $95.35$ & $95.61$ & $96.86$ & $95.6$  & $96.18$ \\
                  & LayoutLMv3   & $96.91$ & $96.23$ & $96.5$  & $96.7$  & $96.43$ & $96.15$ & $96.74$ & $95.76$ & $95.96$ \\
                  & SciBERT      & $96.12$ & $94.96$ & $96.69$ & $96.57$ & $95.54$ & $96.55$ & $96.74$ & $96.13$ & $96.4$  \\
\specialrule{2pt}{1pt}{1pt}
\end{tabular}
\caption{Model performance for each random seed and few-shot episode (Ep), \namelowdiversity~training, after few-shot fine-tuning}\label{tab:lowdiversity_fewshot}
\end{table*}

\begin{table*}
\tiny
\begin{tabular}{lllllllllll}
\specialrule{2pt}{1pt}{1pt}
Test Publisher   & Base Model   &   Seed 0  Ep 0 & Seed 0  Ep 1 & Seed 0  Ep 2 &   Seed 1  Ep 0 & Seed 1  Ep 1 & Seed 1  Ep 2 &   Seed 2  Ep 0 & Seed 2  Ep 1 & Seed 2  Ep 2 \\
\specialrule{1pt}{1pt}{1pt}
\multirow{5}{*}{\textsc{Acta}} & BERT     & $82.88$         & $84.03$         & $81.67$         & $80.18$         & $79.9$          & $80.53$         & $80.28$         & $79.
06$         & $79.37$         \\
                  & LayoutLM     & $81.29$         & $79.49$         & $79.44$         & $82.36$         & $80.96$         & $80.92$         & $81.07$         & $78.
28$         & $80.12$         \\
                  & LayoutLMv2   & $84.51$         & $84.27$         & $84.23$         & $81.58$         & $83.99$         & $83.47$         & $81.52$         & $84.
46$         & $84.03$         \\
                  & LayoutLMv3   & $81.39$         & $81.96$         & $83.05$         & $81.47$         & $81.41$         & $81.27$         & $81.85$         & $84.
68$         & $80.46$         \\
                  & SciBERT      & $83.78$         & $82.18$         & $83.42$         & $83.32$         & $82.14$         & $82.1$          & $80.78$         & $80.
63$         & $79.7$          \\
\hline
\multirow{5}{*}{\textsc{BMC}} & BERT     & $94.76$         & $94.67$         & $94.72$         & $94.67$         & $94.38$         & $94.23$         & $95.44$         & $94.
79$         & $95.14$         \\
                  & LayoutLM     & $94.83$         & $94.79$         & $94.43$         & $95.24$         & $94.8$          & $95.46$         & $95.05$         & $95.
08$         & $95.19$         \\
                  & LayoutLMv2   & $95.19$         & $95.24$         & $95.54$         & $95.64$         & $95.14$         & $95.58$         & $95.41$         & $94.
02$         & $94.99$         \\
                  & LayoutLMv3   & $94.84$         & $95.1$          & $95.24$         & $95.25$         & $94.91$         & $95.74$         & $94.76$         & $94.
22$         & $95.34$         \\
                  & SciBERT      & $95.3$          & $94.12$         & $95.12$         & $95.27$         & $94.52$         & $95.52$         & $94.63$         & $94.
37$         & $95.33$         \\
\hline
\multirow{5}{*}{\textsc{RU}} & BERT    & $92.92$         & $93.29$         & $92.81$         & $92.92$         & $92.99$         & $93.02$         & $93.38$         & $93.
41$         & $92.42$         \\
                  & LayoutLM     & $93.05$         & $93.72$         & $92.92$         & $93.85$         & $94.35$         & $93.36$         & $93.28$         & $93.
8$          & $91.86$         \\
                  & LayoutLMv2   & $94.1$          & $93.92$         & $93.61$         & $93.49$         & $94.04$         & $93.5$          & $93.26$         & $93.
71$         & $92.9$          \\
                  & LayoutLMv3   & $93.34$         & $93.83$         & $93.25$         & $93.8$          & $93.58$         & $93.31$         & $93.75$         & $94.
02$         & $93.85$         \\
                  & SciBERT      & $94.37$         & $93.95$         & $93.51$         & $94.0$          & $93.2$          & $93.8$          & $93.72$         & $94.
06$         & $93.72$         \\
\hline
\multirow{5}{*}{\textsc{PLoS}} & BERT     & $96.51$         & $96.18$         & $96.49$         & $96.34$         & $95.9$          & $95.47$         & $96.47$         & $96.
43$         & $96.52$         \\
                  & LayoutLM     & $96.94$         & $96.09$         & $96.11$         & $97.03$         & $96.89$         & $96.29$         & $96.75$         & $96.
4$          & $96.85$         \\
                  & LayoutLMv2   & $96.77$         & $96.08$         & $95.4$          & $97.02$         & $95.86$         & $95.25$         & $96.77$         & $96.
03$         & $94.93$         \\
                  & LayoutLMv3   & $97.61$         & $96.6$          & $97.06$         & $97.1$          & $96.76$         & $97.05$         & $97.42$         & $96.
43$         & $96.22$         \\
                  & SciBERT      & $97.09$         & $96.36$         & $96.97$         & $96.88$         & $96.73$         & $96.97$         & $96.88$         & $96.
53$         & $96.84$         \\ 
\specialrule{2pt}{1pt}{1pt}
\end{tabular}
\caption{Model performance for each random seed and few-shot episode (Ep), \namemeddiversity~training, after few-shot fine-tuning}\label{tab:meddiversity_fewshot}
\end{table*}

\begin{table*}
\tiny
\begin{tabular}{lllllllllll}
\specialrule{2pt}{1pt}{1pt}
Test Publisher   & Base Model   &   Seed 0  Ep 0 & Seed 0  Ep 1 & Seed 0  Ep 2 &   Seed 1  Ep 0 & Seed 1  Ep 1 & Seed 1  Ep 2 &   Seed 2  Ep 0 & Seed 2  Ep 1 & Seed 2  Ep 2 \\
\specialrule{1pt}{1pt}{1pt}
\multirow{5}{*}{\textsc{Acta}} & BERT      & $79.79$         & $80.22$         & $81.14$         & $79.9$          & $79.41$         & $78.29$         & $80.48$         & $81.22$         & $79.48$         \\
                  & LayoutLM     & $82.53$         & $80.18$         & $80.65$         & $80.6$          & $79.25$         & $80.95$         & $80.87$         & $79.53$         & $79.16$         \\
                  & LayoutLMv2   & $81.27$         & $82.57$         & $83.29$         & $81.24$         & $82.54$         & $80.56$         & $81.2$          & $83.69$         & $83.68$         \\
                  & LayoutLMv3   & $81.53$         & $82.74$         & $81.06$         & $82.73$         & $82.67$         & $80.9$          & $81.44$         & $82.71$         & $81.14$         \\
                  & SciBERT      & $81.81$         & $82.98$         & $82.37$         & $81.8$          & $82.17$         & $82.74$         & $80.64$         & $82.03$         & $81.26$         \\ 
\hline
\multirow{5}{*}{\textsc{BMC}} & BERT      & $95.21$         & $94.97$         & $94.97$         & $95.58$         & $94.88$         & $95.2$          & $95.65$         & $95.08$         & $95.33$         \\
                  & LayoutLM     & $94.93$         & $95.34$         & $95.38$         & $95.11$         & $94.55$         & $94.86$         & $93.82$         & $94.12$         & $94.78$         \\
                  & LayoutLMv2   & $95.04$         & $93.33$         & $95.47$         & $95.09$         & $94.94$         & $94.82$         & $94.63$         & $94.68$         & $94.66$         \\
                  & LayoutLMv3   & $95.57$         & $94.54$         & $94.58$         & $95.22$         & $93.77$         & $94.67$         & $95.03$         & $93.2$          & $94.02$         \\
                  & SciBERT      & $94.86$         & $94.15$         & $94.56$         & $95.32$         & $94.82$         & $95.07$         & $94.98$         & $93.34$         & $94.66$         \\
\hline
\multirow{5}{*}{\textsc{RU}} & BERT     & $93.36$         & $93.27$         & $92.78$         & $93.39$         & $93.55$         & $93.03$         & $93.48$         & $93.67$         & $93.65$         \\
                  & LayoutLM     & $94.23$         & $93.75$         & $92.89$         & $93.96$         & $93.8$          & $92.71$         & $93.86$         & $93.97$         & $93.06$         \\
                  & LayoutLMv2   & $93.95$         & $93.56$         & $92.86$         & $94.31$         & $93.68$         & $93.18$         & $94.04$         & $93.67$         & $93.33$         \\
                  & LayoutLMv3   & $93.94$         & $94.73$         & $93.8$          & $94.04$         & $94.07$         & $93.89$         & $94.4$          & $93.97$         & $93.74$         \\
                  & SciBERT      & $94.03$         & $93.09$         & $92.12$         & $93.74$         & $93.2$          & $93.48$         & $94.0$          & $93.97$         & $93.35$         \\
\hline
\multirow{5}{*}{\textsc{PLoS}} & BERT      & $96.2$          & $95.5$          & $95.81$         & $96.41$         & $95.66$         & $95.36$         & $96.54$         & $96.6$          & $96.47$         \\
                  & LayoutLM     & $96.92$         & $97.02$         & $97.07$         & $96.57$         & $96.34$         & $96.22$         & $97.2$          & $96.52$         & $96.51$         \\
                  & LayoutLMv2   & $96.42$         & $95.01$         & $95.13$         & $96.7$          & $96.19$         & $95.71$         & $97.19$         & $95.84$         & $95.01$         \\
                  & LayoutLMv3   & $97.03$         & $96.27$         & $96.25$         & $96.53$         & $96.06$         & $96.37$         & $96.79$         & $95.7$          & $95.66$         \\
                  & SciBERT      & $97.06$         & $96.3$          & $96.58$         & $97.3$          & $96.76$         & $97.12$         & $96.95$         & $96.42$         & $96.76$         \\
\specialrule{2pt}{1pt}{1pt}
\end{tabular}
\caption{Model performance for each random seed and few-shot episode (Ep), \namehighdiversity~training, after few-shot fine-tuning}\label{tab:highdiversity_fewshot}
\end{table*}
\end{document}